\documentclass[runningheads]{llncs}
\usepackage{graphicx}

\usepackage{tikz}
\usepackage{comment}
\usepackage{amsmath,amssymb} 
\usepackage{adjustbox}
\usepackage{color}
\usepackage{pifont}  
\usepackage{booktabs}  
\usepackage{multirow}
\usepackage{caption}
\usepackage{subcaption}

\newcommand{\cmark}{\ding{51}}%
\newcommand{\xmark}{{\color{black} \ding{55}}}%

\usepackage{hyperref}

\begin{document}
\pagestyle{headings}
\mainmatter
\def\ECCVSubNumber{***}  

\title{OCR-IDL: OCR Annotations for Industry Document Library Dataset} 

\newcommand{\DatasetName}{XXX}
\titlerunning{OCR-IDL}
%
\author{Ali Furkan Biten\inst{1} \and
Rub{\`e}n Tito\inst{1} \and
Lluis Gomez\inst{1} \and
Ernest Valveny\inst{1} \and
Dimosthenis Karatzas\inst{1}}
\authorrunning{Biten et al.}
%
\institute{Computer Vision Center, UAB \and
\email{\{abiten,rperez,lgomez,ernest,dimos\}@cvc.uab.cat}}
\maketitle

\begin{abstract}
Pretraining has proven successful in Document Intelligence tasks where deluge of documents are used to pretrain the models only later to be finetuned on downstream tasks. One of the problems of the pretraining approaches is the inconsistent usage of pretraining data with different OCR engines leading to incomparable results between models. In other words, it is not obvious whether the performance gain is coming from diverse usage of amount of data and distinct OCR engines or from the proposed models. To remedy the problem, we make public the OCR annotations for IDL documents using commercial OCR engine given their superior performance over open source OCR models. The contributed dataset (OCR-IDL) has an estimated monetary value over 20K US\$. It is our hope that OCR-IDL can be a starting point for future works on Document Intelligence. All of our data and its collection process with the annotations can be found in \url{https://github.com/furkanbiten/idl\_data}.
\end{abstract}

\section{Introduction}
Analysis of masses of scanned documents  is  essential  in  intelligence, law, knowledge  management, historical scholarship,  and other areas~\cite{lewis2006building}. The documents are often complex and varied in nature that can be digital or scanned born, containing elements such as forms, figures, tables, graphics and photos, while being produced by various printing and handwriting technologies. Some common examples of documents comprise of purchase orders, financial reports, business emails, sales agreements, vendor contracts, letters, invoices, receipts, resumes, and many others~\cite{xu2020layoutlm}. Processing various document types to user's intent is done with manual labor that is time-consuming and expensive, meanwhile requiring manual customization or configuration.
In other words, each type of document demands hard-coded changes when there is a slight change in the rules or workflows of documents or even when dealing with multiple formats.

To address these problems, Document Intelligence models and algorithms are created to automatically structure, classify and extract information from documents, improving automated document processing. Particularly, Document Intelligence as a research field aims at creating models for automatically analyzing and understanding documents, reducing the time and the cost associated with it. From a research perspective, what makes Document Intelligence especially challenging is the requirement of combining various disciplines such as optical character recognition (OCR), document structure analysis, named entity recognition, information retrieval, authorship attribution and many more.

Recent methods on Documents Intelligence utilize deep neural networks combining Computer Vision and Natural Language Processing. Hao \textit{et al.}~\cite{hao2016table} proposed an end-to-end training using Convolutional Neural Networks to detect tables in documents. Several published works~\cite{soto2019visual,schreiber2017deepdesrt,zhong2019publaynet} exploit the advances in object detection~\cite{ren2015faster,he2017mask} to further improve the accuracy in document layout analysis. Even though these works have advanced the Document Intelligence field, there are two main limitations to be recognized: (i) they rely on a small human annotated dataset and (ii) they use pre-trained networks that have never seen any documents, hence the interaction between text and layout. Inspired by BERT~\cite{devlin2018bert},Xu \textit{et al.}~\cite{xu2020layoutlm} identified these problems and propose a pre-training strategy to unlock the potential of large-scale unlabeled documents. More specifically, they obtain OCR annotations from an open source OCR engine Tesseract~\cite{smith2007overview} for 5 Million documents from IIT-CDIP~\cite{lewis2006building} dataset. With the introduction of pre-training strategy and advances in modern OCR engine~\cite{aberdam2021sequence,fang2021read,he2021visual,litman2020scatter,na2021multi}, many contemporary approaches~\cite{biten2021latr,appalaraju2021docformer,xu2020layoutlmv2} have utilized even more data to advance the Document Intelligence field.

In this work, we make public the OCR annotations for 26 Millions pages using a commercial OCR engine that has the monetary value over 20K US\$. Our motivation for releasing a massive scale documents dataset annotated with a commercial OCR engine is two-fold. First of all, the usage of different amount of documents and different OCR engines across the papers makes it impossible to fairly compare their results and hence their architecture. By creating this dataset, we hope that the works in Document Intelligence will become more comparable and have better intuition on what the proposed architecture can actually accomplish. 

Secondly, we decide to use a commercial OCR engine, specifically Amazon Textract\footnote{\url{https://aws.amazon.com/textract/}}, over Tesseract. It is because the performance of the OCR engines can significantly affect the model's performance which can be seen in fields that use OCR annotations, such as in fine-grained classification~\cite{mafla2020fine,mafla2021multi,mafla2021stacmr}, in scene-text visual question answering~\cite{biten2019scene,singh2019towards,biten2019icdar,gomez2021multimodal}, in document visual question answering (DocVQA)~\cite{tito2021document,mathew2021docvqa}. Apart from improving the annotation quality significantly, we want to level the differences between research groups and companies.

\begin{figure}[ht!]
\begin{adjustbox}{max width=1.5\textwidth,center}
     \centering
     \begin{subfigure}[b]{0.25\textwidth}
         \centering
         \includegraphics[width=\textwidth]{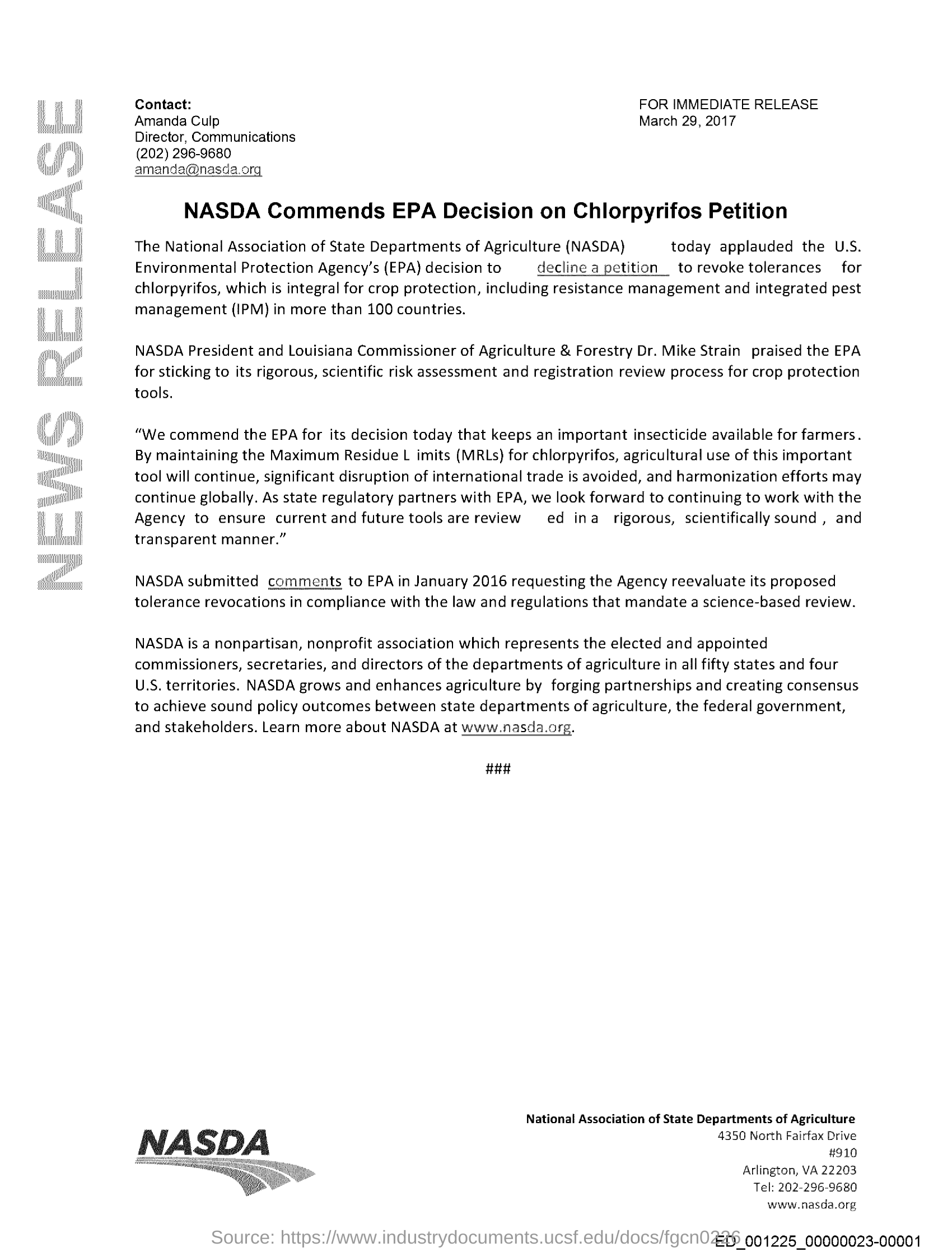}
         \caption{}
         \label{fig:qual1}
     \end{subfigure}
     \hfill
     \begin{subfigure}[b]{0.25\textwidth}
         \centering
         \includegraphics[width=\textwidth]{media/images/qualitatives/khnk0226_4.png}
         \caption{}
         \label{fig:qual2}
     \end{subfigure}
     \hfill
     \begin{subfigure}[b]{0.25\textwidth}
         \centering
         \includegraphics[width=\textwidth]{media/images/qualitatives/fggn0226_77.png}
         \caption{}
         \label{fig:qual3}
     \end{subfigure}
     \begin{subfigure}[b]{0.25\textwidth}
         \centering
         \includegraphics[width=\textwidth]{media/images/qualitatives/flmh0077_12.png}
         \caption{}
         \label{fig:qual4}
     \end{subfigure}
     \begin{subfigure}[b]{0.25\textwidth}
         \centering
         \includegraphics[width=\textwidth]{media/images/qualitatives/fgfl0228_3.png}
         \caption{}
         \label{fig:qual5}
     \end{subfigure}
    \end{adjustbox}
\caption{\textbf{Document images of OCR-IDL.} The dataset includes a wide variety of documents with dense text (a), tables (b), figures (c), and complex layouts that combines different elements (d, e).}
\label{fig:qualitative}
\end{figure}


We provide the annotations for publicly available documents from Industry Documents Library (IDL). IDL is a digital archive of documents created by industries which influence public health, hosted by the University of California, San Francisco Library~\footnote{\url{https://www.industrydocuments.ucsf.edu}}. IDL has already been used in the literature for building datasets: IIT-CDIP~\cite{lewis2006building}, RVL-CDIP~\cite{harley2015evaluation}, DocVQA~\cite{tito2021document,mathew2021docvqa}. Hence, our OCR annotations can be used to further advance in these tasks. 

The rest of the paper is structured as follows. First, we briefly explain all the related works. Next, we will elaborate on our data collection and comparison to other datasets. Finally, we will provide various statistics of the annotations and conclude our paper. 

\section{Related Work}
Document Intelligence can be considered as an umbrella term covering problems of Key Information Extraction \cite{carbonell2021named,xu2020layoutlm}, Table Detection \cite{riba2019table,prasad2020cascadetabnet} and Structure Recognition \cite{raja2020table,zhong2020image}, Document Layout Segmentation \cite{biswas2021beyond,biswas2022docsegtr} Document Layout Generation \cite{biswas2021docsynth,patil2020read,arroyo2021variational,souibgui2022one}, Document Visual Question Answering \cite{tito2021icdar,tito2021document,mathew2022infographicvqa}, Document Image Enhancement \cite{souibgui2020gan,jemni2022enhance,souibgui2022docentr} which involves the understanding of visually rich semantic information and structure of different layout entities of a whole page.

Early days of Document Intelligence has relied on rule-based handcrafted approaches mainly divided into bottom-up and top-down methods. Bottom-up methods~\cite{ha1995document,lebourgeois1992fast,o1993document,simon1997fast} first detect connected components at the pixel level, later to be fused into higher level of structure through various heuristics and name depending on distinct structural features. While top-down methods~\cite{ha1995recursive} dissects a page into smaller units such as titles, text blocks, lines, and words.

Lately, the success of large-scale pre-training~\cite{devlin2018bert} in Natural Language Processing has been integrated into Document Intelligence, resulting in impressive performance gains. These methods follow a two step procedure where first they pretrain the models on unlabeled documents (OCR annotations are obtained by an off-the-shelf OCR engine), then they finetune it on specific downstream tasks. LayoutLM~\cite{xu2020layoutlm} is one of the first works that pretrain BERT based language model with document layout information, using masked language/vision modeling and multi label classification. BROS~\cite{hong2020bros} is built on top of Span-BERT~\cite{joshi2020spanbert} with spatially aware graph decoder. For the pretraining loss, they use area-masked language model. Self-Doc~\cite{li2021selfdoc} utilizes two separate transformer~\cite{vaswani2017attention} encoders for visual and textual features and later to be fed to multi-modal transformers encoder. TILT~\cite{powalski2021going} tries to encode the layout information by integrating pairwise 1D and 2D information into their models. Uni-Doc~\cite{gu2021unidoc} is designed to do most document understanding tasks that takes words and visual features from a semantic region of a document image by combining three self-supervised losses. More recent methods, Doc-Former~\cite{appalaraju2021docformer} and LayoutLMv2~\cite{xu2020layoutlmv2} combine multiple pretraining losses such as image-text alignment, learning to construct image features and multi-modal masked language modeling together to achieve state-of-the-art results. 

Yet, comparing all of these works are cumbersome since each work that performs pretraining uses different amounts of data with diverse OCR engines. Hence, this makes it especially hard to understand where the gain is coming from. In other words, we can not draw clear conclusions to questions such as: ``What is the effect of the amount of pretraining data on the performance?'', ``Is the performance gain coming from a better/stronger OCR engine or from the proposed architecture?'', ``What is the effect of the pretraining loss on the downstream tasks keeping OCR and the amount of data identical?'' To help answer these questions, we collect and annotate the largest public OCR annotated documents dataset (OCR-IDL). 

\section{OCR-IDL Dataset}
In this section, we elaborate on various details regarding OCR-IDL. Firstly, we explain the process we follow on how we get the IDL data and use Amazon-Textract to obtain OCR annotations. Next, we compare OCR-IDL to other datasets that have proven useful for the document intelligence tasks. And finally, we provide in-depth statistics on the documents we use. 

\newcommand{\mydagger}{\textsuperscript{\textdagger}}

\begin{table}[t!]
\begin{adjustbox}{max width=1.5\textwidth,center}
\begin{tabular}{lccllcccc}
\toprule
Dataset                               & \# of Docs       & \# of Pages & Docs source  & Docs. description     & OCR-Text        & OCR-BB & Layout & Doc. type\\
\midrule
IIT-CDIP~\cite{lewis2006building}     & $6.5M$*         & $35.5M$*  & UCSF-LTD      & Industry documents    & Unknown         & \xmark & \xmark & \cmark \\
RVL-CDIP~\cite{harley2015evaluation}  & -\mydagger      & $400K$    & UCSF-LTD      & Industry documents    & \xmark          & \xmark & \xmark & \cmark \\
PublayNet~\cite{zhong2019publaynet}   & -\mydagger      & $364K$    & PubMedCentral & Journals and articles & \xmark          & \xmark & \cmark & \xmark \\
DocBank~\cite{li2020docbank}          & -\mydagger      & $500K$    & arXiv         & Journals and articles & \xmark          & \xmark & \cmark & \xmark \\
DocVQA~\cite{tito2021icdar}           & $6K$            & $12K$     & UCSF-IDL      & Industry documents    & Microsoft OCR   & \cmark & \xmark & \cmark \\
OCR-IDL                               & $4.6M$          & $26M$     & UCSF-IDL      & Industry documents    & Amazon Textract & \cmark & \xmark & \cmark \\
\bottomrule
\end{tabular}
\end{adjustbox}
\vspace{0.1cm}
\caption{Summary of other Document Intelligence Datasets. *We skipped $145K$ documents that gave xml parsing errors, didn't contain document ID or number of pages. \textsuperscript{\textdagger}No traceability between different pages of the same document.}
\label{tab:dataset_comp_ext}
\end{table}
\subsection{Data Collection}
As already mentioned, IDL is an industry documents library hosted by UCSF, its main purpose is to ``identify, collect, curate, preserve, and make freely accessible internal documents created by industries and their partners which have an impact on public health, for the benefit and use of researchers, clinicians, educators, students, policymakers, media, and the general public at UCSF and internationally''\footnote{\url{https://en.wikipedia.org/wiki/Industry\_Documents\_Library}}. IDL in total contains over 70 millions documents. We use the publicly available link\footnote{\url{https://s3-us-west-2.amazonaws.com/edu.ucsf.industrydocuments.artifacts/}} to downlad the 4.6 Million Documents from IDL dataset which comes in the format of PDFs. Some examples can be viewed in Figure~\ref{fig:qualitative}. We appreciate that the documents are quite varied in terms of layout where there are tables, figures, ads and combination of them in a single page. Also, we see that the documents are quite text-rich containing many OCR words. Finally, the pages contain smudges, taints and other discoloration what is found in the in a real use-case scenario. 

\begin{figure*}[t!]
\begin{adjustbox}{max width=2.\textwidth,center}
    \centering
    \includegraphics[width=1.1\textwidth]{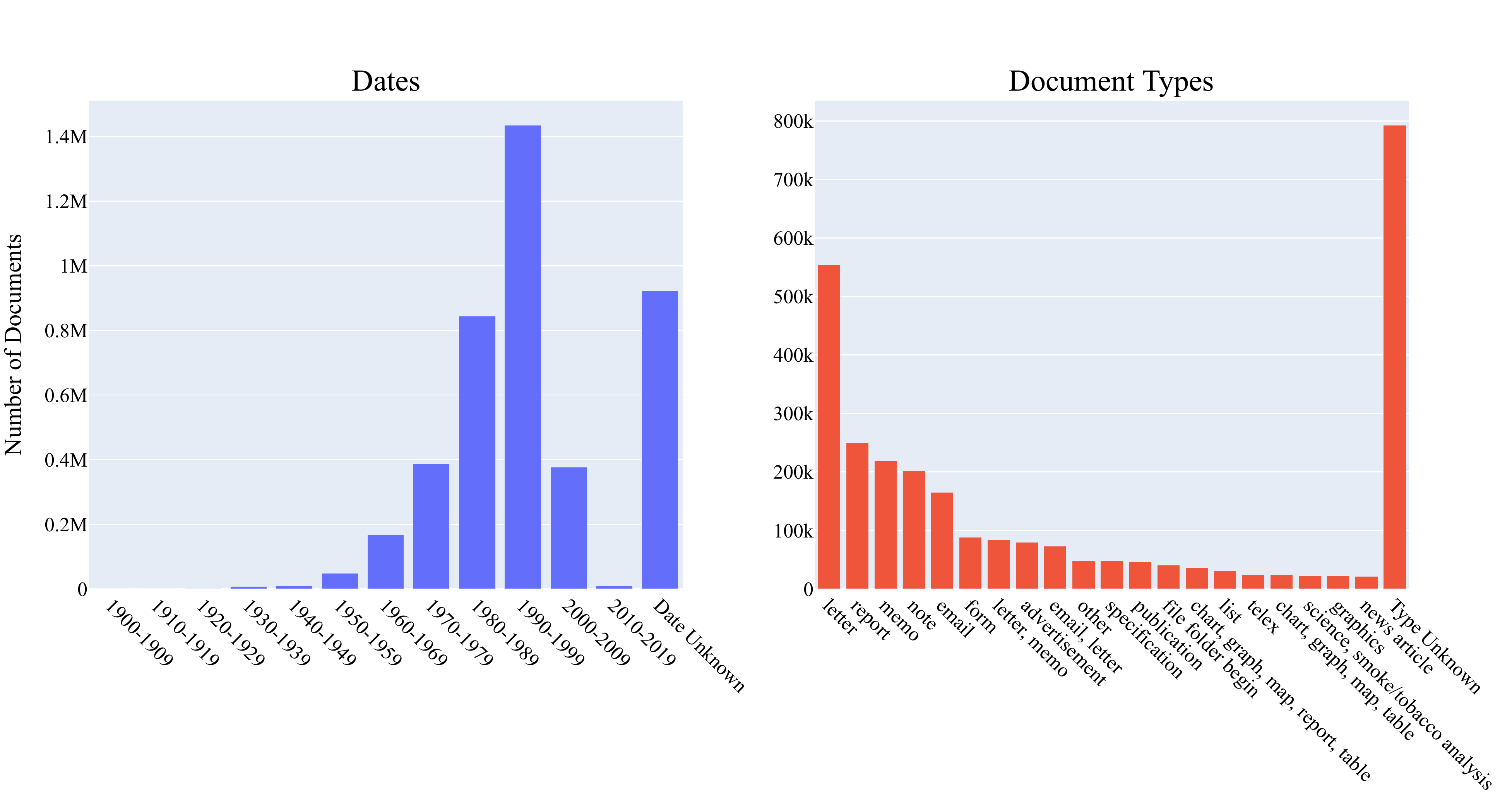}
    \end{adjustbox}
    \caption{Distribution of the annotated documents in terms of document types and dates.}
    \label{fig:typesVSdate}
\end{figure*}
We choose to annotate only 4.6M documents since annotating 13M documents not only would be much costlier (42K dollars instead of 18K) but also in the literature it is shown that using more data have diminishing returns on the downstream task~\cite{biten2021latr}. After obtaining the data, we preprocess the documents to remove empty, faulty and broken pdfs. This process resulted in the elimination of 6548 documents. Moreover, we remove also documents that have more than 2900 pages which are 71 in total. The necessity of removing such huge documents was because Amazon-Textract OCR only accepts up to 3000 pages. After pre-processing the documents, we feed all the documents to the OCR engine to obtain the annotations. The annotations provided by the Textract engine include transcription of words and lines and their corresponding bounding boxes and polygons with text type that can be printed or handwritten. Processing all 4.6M documents was done by a single machine with 16 parallelized cores and took about 1 month.

\begin{figure*}[t!]
\begin{adjustbox}{max width=2.\textwidth,center}
    \centering
    \includegraphics[width=1.1\textwidth]{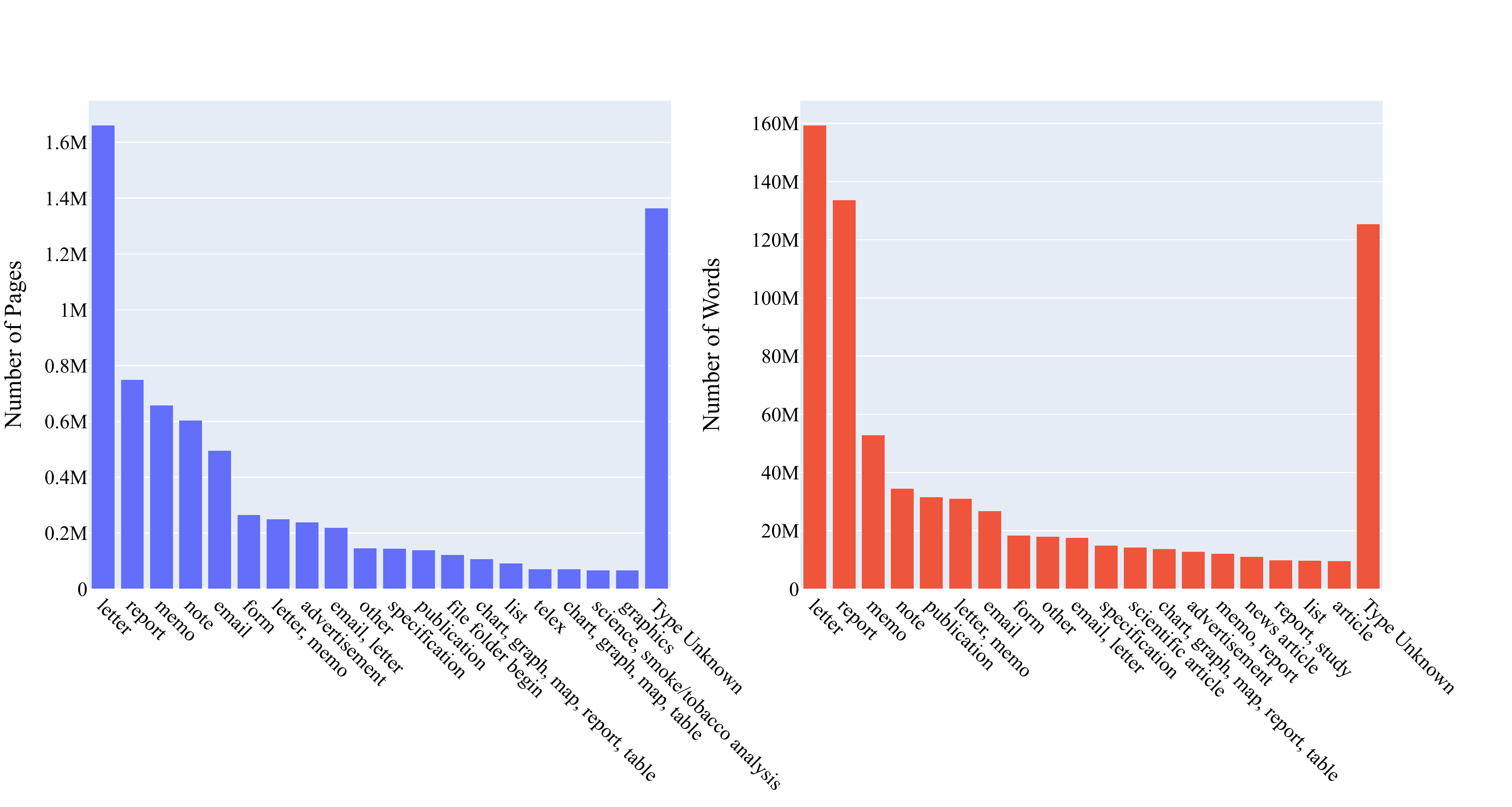}
    \end{adjustbox}
    \caption{Distribution of number of pages and number of words per document type.}
    \label{fig:page_words_by_type}
\end{figure*}
\subsection{Comparison to existing datasets}
In this section, we compare the statistics of the amount of documents and pages to other datasets that are used in Document Intelligence. To name a few, the Illinois Institute of Technology dataset for Complex Document Information Processing (IIT-CDIP)~\cite{lewis2006building} is the biggest document dataset and it is designed for the task of information retrieval. The Ryerson Vision Lab Complex Document Information Processing (RVL-CDIP)~\cite{harley2015evaluation} dataset used the IIT-CDIP metadata to create a new dataset for document classification. PublayNet~\cite{zhong2019publaynet} and DocBank~\cite{li2020docbank} are datasets designed for layout analysis tasks and DocVQA~\cite{mathew2021docvqa,tito2021icdar} instead, is designed for Visual Question Answering task over document images.

We summarize all the key information for comparison in Table~\ref{tab:dataset_comp_ext}. First, we stress that OCR-IDL is the second biggest dataset in amount for pre-training and biggest dataset with annotations obtained from commercial OCR. This provides unique opportunity for the Document Intelligence research for utilizing the unlabeled documents in their research. Furthermore, even though OCR-IDL uses the documents from the same source as IIT-CDIP and RVL-CDIP, it also contains other type of industrial documents. OCR-IDL contains documents from chemical, medical and drug industries, hence having more variety in terms of content as well as layout information.

\begin{figure*}[t!]
\begin{adjustbox}{max width=2.\textwidth,center}
    \centering
    \includegraphics[width=1.1\textwidth]{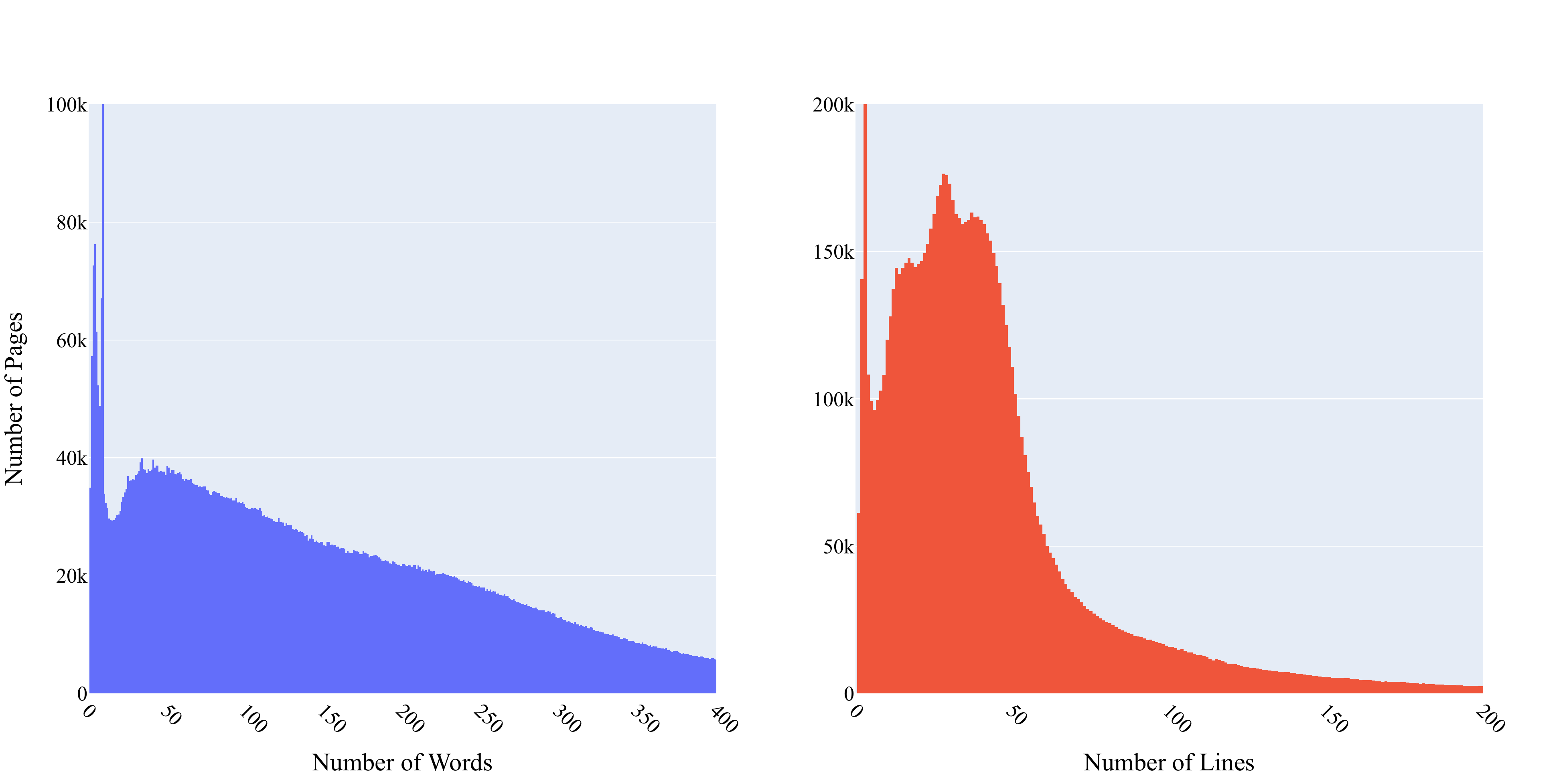}
    \end{adjustbox}
    \caption{Distribution of number of words (left) and lines (right) by pages.}
    \label{fig:word_line_by_pages}
\end{figure*}
\subsection{Dataset Statistics}
IDL documents come with metadata that is curated by human annotators. They include information about the date, industry, drugs, chemicals, document types and many more. 
We restrict our analysis on exploring what type of documents we have and the distribution of the dates they are created which can be found in Figure~\ref{fig:typesVSdate}. In IDL metadata, there are 35k various document types from which we show only the most common 20 which include letters, report, email, memo, note, etc. As can be seen in the left side of Figure~\ref{fig:typesVSdate}, most of the documents' type is unknown. Moreover, even though we have a skewed distribution on letter, report and memo, we also have very distinct distribution such as chart, graphics, news articles. This is especially encouraging because it provides diverse layouts within various contexts. Moreover, the documents are created spanning 100 years where most of the documents are in the range of 1980-2010 as can be seen in the right side of Figure~\ref{fig:typesVSdate}. The date the documents are created contributes to the variability of the documents not only in terms of semantics but more importantly having visual artifacts (smudges, different resolution) with different printing technologies.

On top of the amount of documents, we give more details on the number of pages and words for each document type. The number of pages follows the same distribution as the amount of documents per type, as can be appreciated in Figure~\ref{fig:page_words_by_type}. Also, we can see from Figure~\ref{fig:page_words_by_type} that the report type is much richer in terms of text while the rest follows more or less the same distribution.


\begin{figure}[t!]
    \centering
    \includegraphics[width=1.\linewidth]{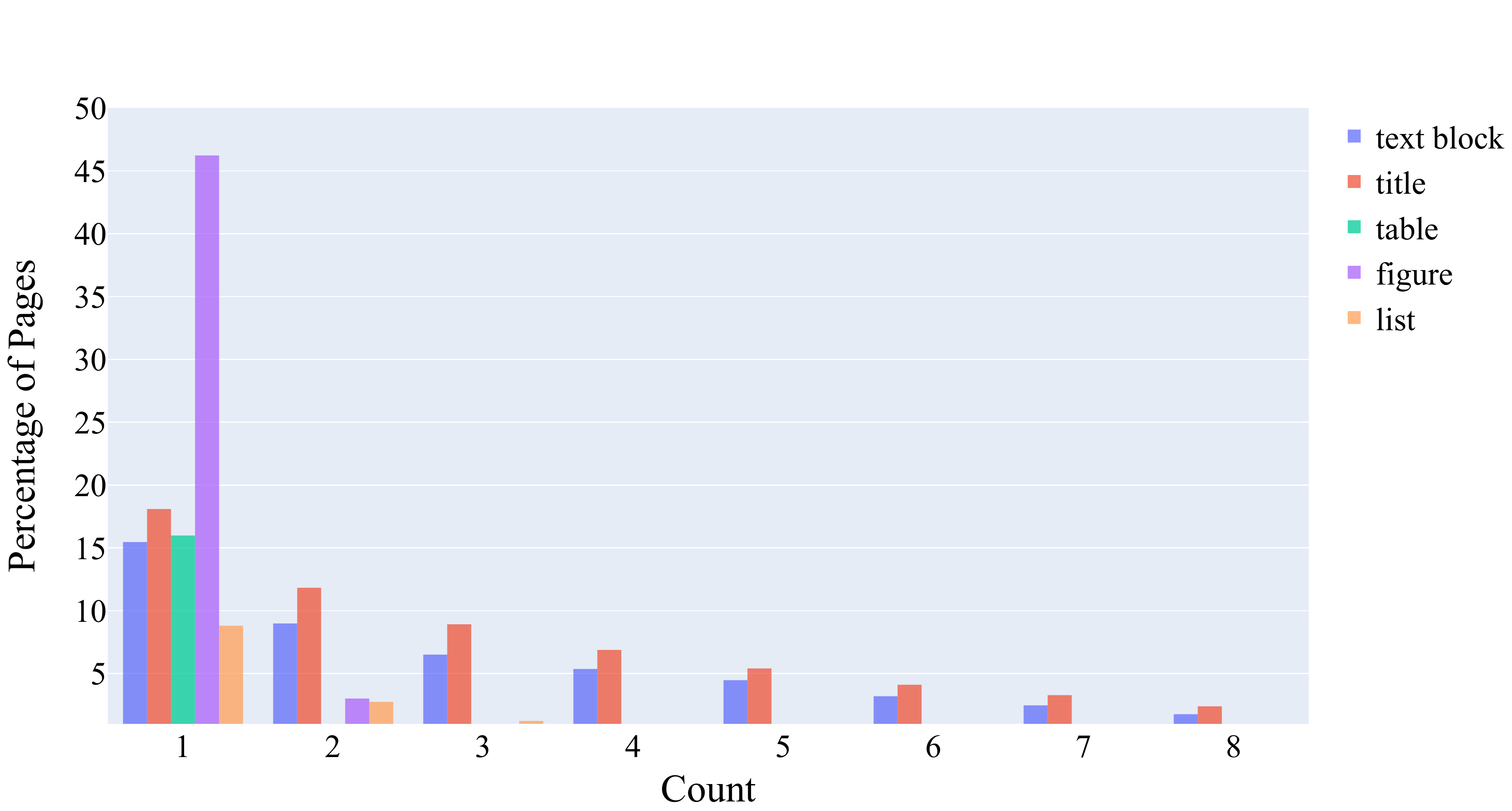}
    \caption{Distribution of various layout blocks in OCR-IDL. Our documents contain a lot of diversity in terms of title, figure, text block, list and table.}
    \label{fig:layout}
\end{figure}

We turn our attention to OCR annotation statistics. In total, we obtain OCR annotations for 4614232 (4.6M) documents where we have 26621635 (26M) pages, averaging 6 pages per document. Moreover, since documents are known to be a text-rich environment, we provide extra details regarding the amount of words and lines per page and documents. We have 166M words with 46M lines, on average there are 62.5 words and 17.5 lines per page while 360.8 words and 101.25 lines per document. To have a better understanding of the distribution of words and lines per page, we present Figure~\ref{fig:word_line_by_pages}. As shown in the figure, mean distribution is between 20 to 100 words per page while there is a significant amount of pages that contain more than 200 words per page. The distribution for lines follows a different distribution in which it can be observed that most of the pages contain from 10 to 50 lines. In either case, it is clearly observed that documents at hand are ideal for performing pretraining with their diverse layouts and text-rich settings.


\begin{figure}[ht!]
     \centering
     \setlength{\tabcolsep}{0.01pt}
     \begin{tabular}{ccccc}
     \includegraphics[width=0.2\textwidth]{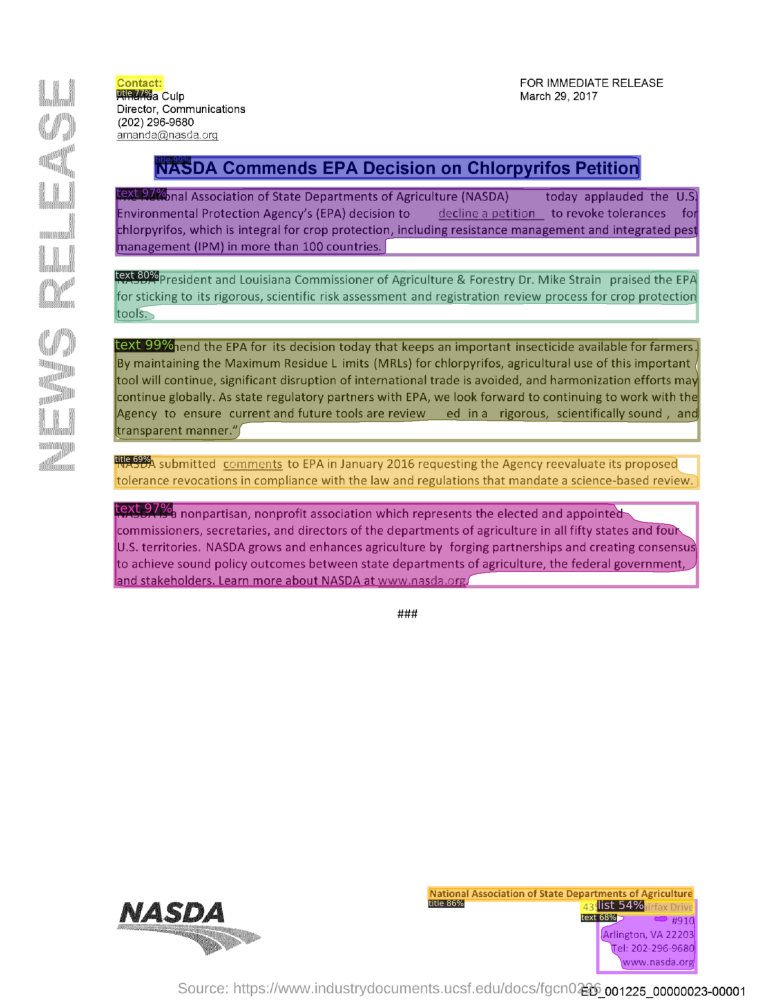}
     \includegraphics[width=0.2\textwidth]{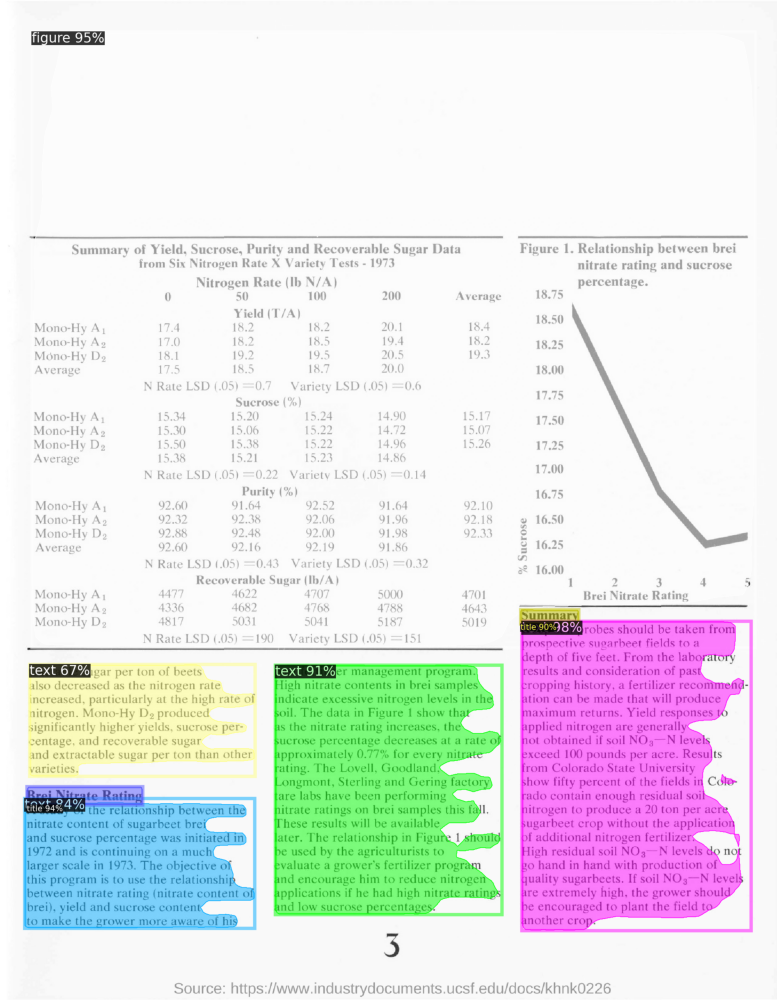}
    \includegraphics[width=0.2\textwidth]{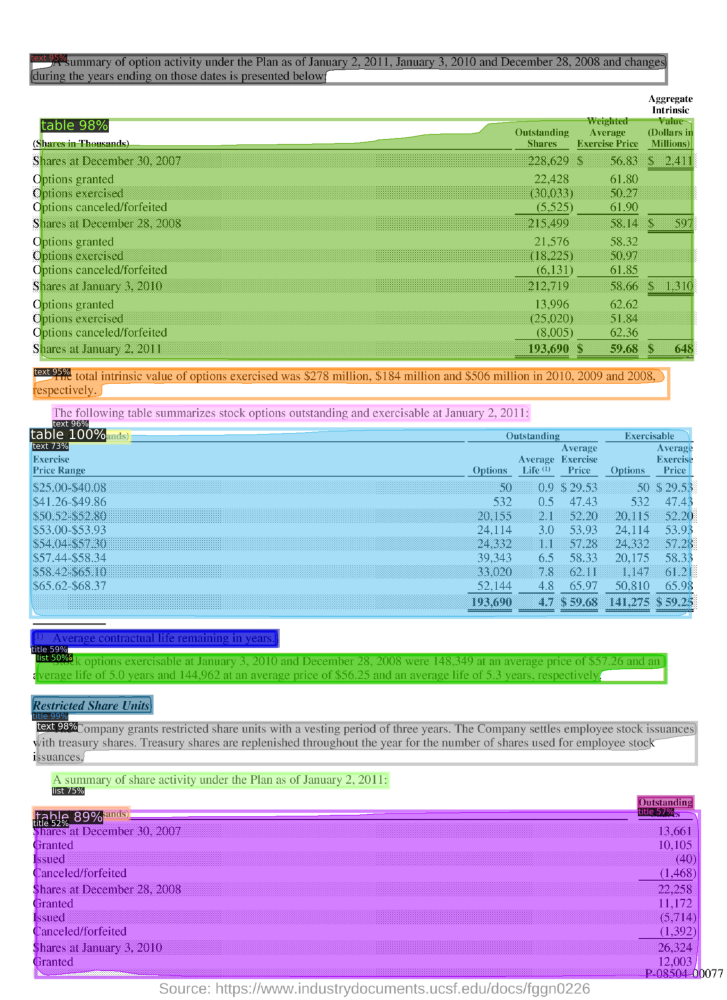}
    \includegraphics[width=0.2\textwidth]{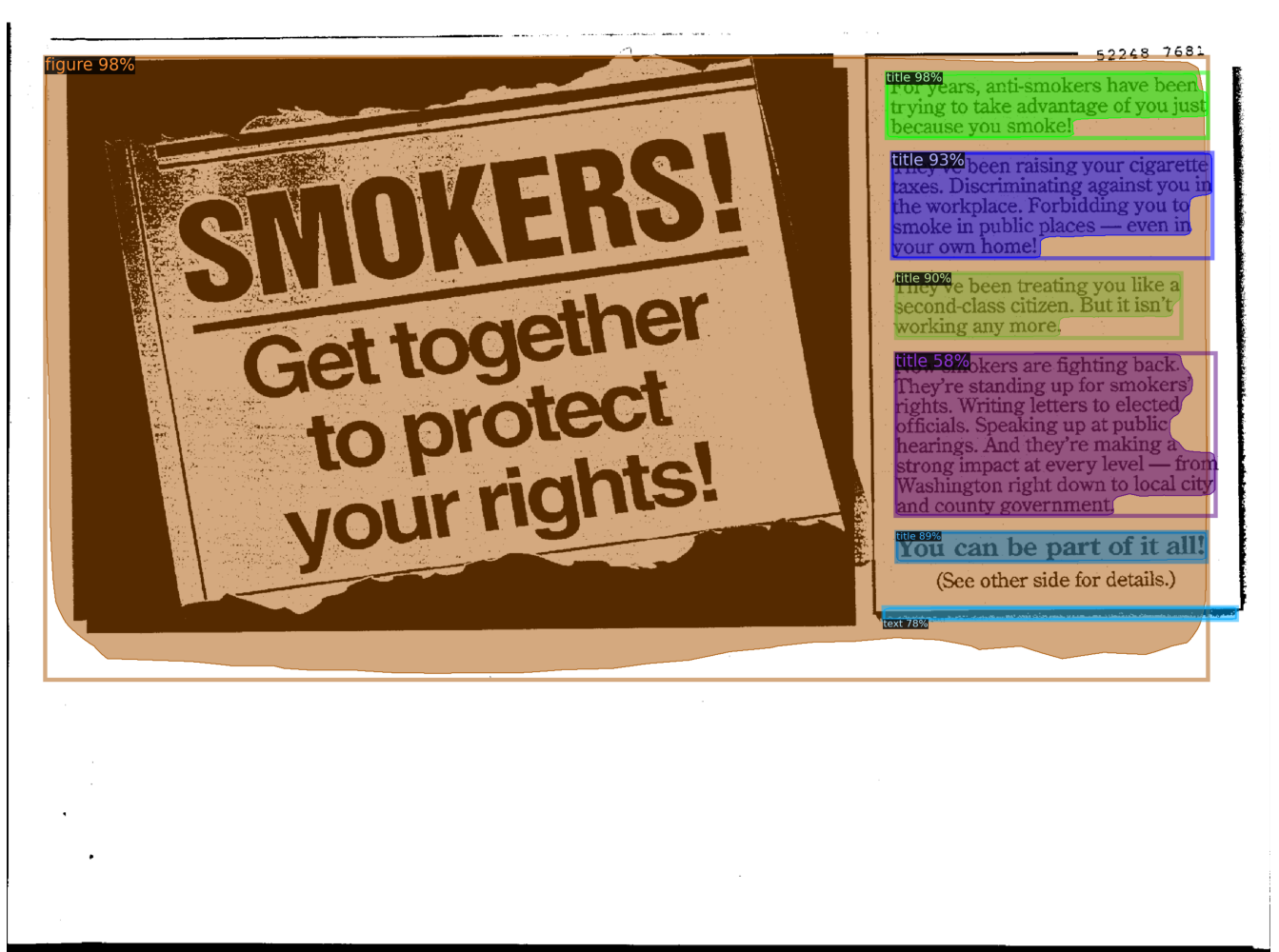}
    \includegraphics[width=0.2\textwidth]{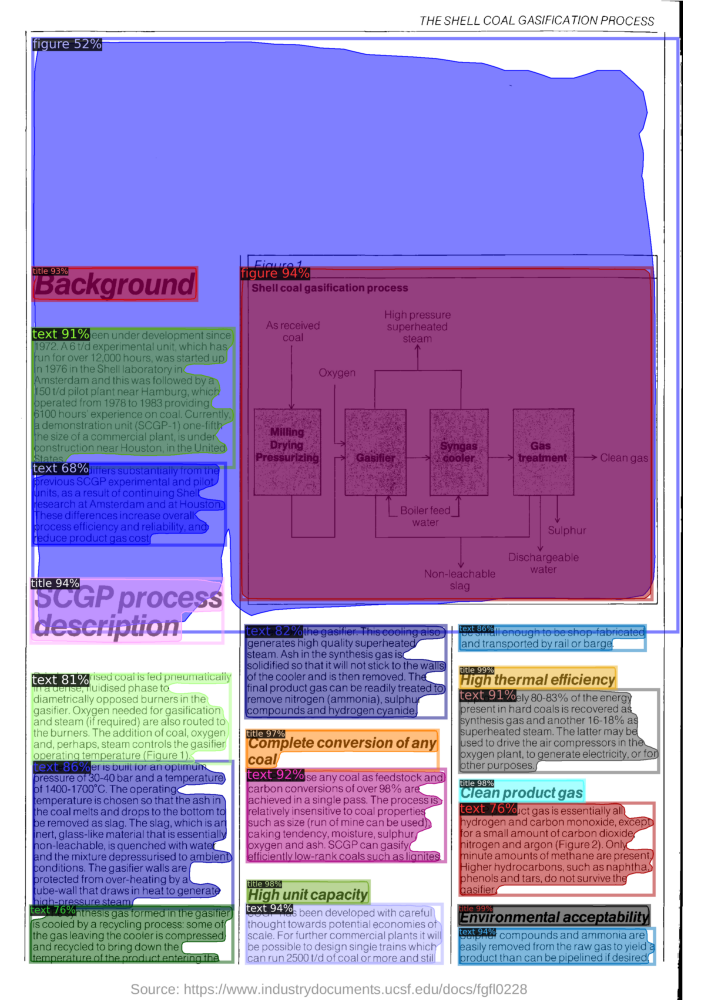}
    \end{tabular}
\caption{{Qualitative results for segmentation of layout information in OCR-IDL.}}
\label{fig:seg_result}
\end{figure}

Finally, to quantify the diversity of the documents in terms of layout, we run publicly available Faster-RCNN~\cite{ren2015faster} trained on PubLayNet~\cite{zhong2019publaynet} to segment a document into text block, title, figure, list, table. To obtain the segmentation results, we randomly selected 40K pages and some segmentation examples can be found in Figure~\ref{fig:seg_result}.  It can be appreciated from Figure~\ref{fig:layout} that 40\% of the documents have at least 1 figure. We also observe that 10-20\% of the pages have at least 1 table and list, showing that documents at hand contain very diverse layout information. Moreover, more than 45\% of the pages contain more than 1 text block and 1 title, making the documents a text rich environment for pre-training on Document Intelligence tasks.

\section{Conclusion}
In this paper, we have presented our effort to provide OCR annotations for the large-scale IDL document dataset called OCR-IDL. These annotations have a monetary value over \$20,000 and are made publicly available with the aim of advancing the Document Intelligence research field.
Our motivation is two-fold, first we make use of a commercial OCR engine to obtain high quality annotations, leading to reduce the noise provided by OCR on pretraining and downstream tasks. Secondly, it is our hope that OCR-IDL can be a starting point for future works on Document Intelligence to be more comparable.
Throughout this article we have detailed the process that we have followed to obtain the annotations, we have presented a statistical analysis, and compared them with other datasets in the state of the art. The provided analysis shows that our contribution has a high potential to be used successfully in pre-training strategies for document intelligence models. All the code for data collection process and annotations can be accessed in \url{https://github.com/furkanbiten/idl\_data}.

%
%
\bibliographystyle{splncs04}
\bibliography{egbib}
\end{document}